\pdfoutput=1

\documentclass[11pt]{article}

\usepackage[]{acl}

\usepackage{times}

\usepackage{latexsym}
\usepackage{booktabs}
\usepackage{amsmath}
\usepackage{multirow}
\usepackage{amssymb}

\makeatletter
\newcommand{\printfnsymbol}[1]{%
  \textsuperscript{\@fnsymbol{#1}}%
}
\makeatother

\usepackage[T1]{fontenc}

\usepackage[utf8]{inputenc}

\usepackage{microtype}

\usepackage{graphicx}
\usepackage{tabularx}
\title{Efficient Cluster-Based $k$-Nearest-Neighbor Machine Translation}

\author{Dexin Wang\thanks{~~~Equal contribution.}\\ Tianjin University \\
        \texttt{dexinwang@tju.edu.cn}
        \And
        Kai Fan\printfnsymbol{1}  \\  \texttt{interfk@gmail.com}
         \AND
        Boxing Chen \\  \texttt{chenboxing@gmail.com}
          \And
        Deyi Xiong\thanks{~~~Corresponding author.}  \\ Tianjin University \\ \texttt{dyxiong@tju.edu.cn}
}

\begin{document}
\maketitle
\begin{abstract}
$k$-Nearest-Neighbor Machine Translation ($k$NN-MT) has been recently proposed as a non-parametric solution for domain adaptation in neural machine translation (NMT).
It aims to alleviate the performance degradation of advanced MT systems in translating out-of-domain sentences by coordinating with an additional token-level feature-based retrieval module constructed from in-domain data.
Previous studies \cite{knnmt,adaptiveknnmt} have already demonstrated that non-parametric NMT is even superior to models fine-tuned on out-of-domain data.
In spite of this success, $k$NN retrieval is at the expense of high latency, in particular for large datastores.
To make it practical, in this paper, we explore a more efficient $k$NN-MT and propose to use clustering to improve the retrieval efficiency.
Concretely, we first propose a cluster-based Compact Network for feature reduction in a contrastive learning manner to compress context features into 90+\% lower dimensional vectors.
We then suggest a cluster-based pruning solution to filter out 10\%\textasciitilde40\% redundant nodes in large datastores while retaining translation quality.
Our proposed methods achieve better or comparable performance while reducing up to 57\% inference latency against the advanced non-parametric MT model on several machine translation benchmarks. 
Experimental results indicate that the proposed methods maintain the most useful information of the original datastore and the Compact Network shows good generalization on unseen domains. 
Codes are available at \url{https://github.com/tjunlp-lab/PCKMT}.
\end{abstract}

\section{Introduction}

Recently, non-parametric approaches \cite{knnmt,adaptiveknnmt,zheng2021,kernal-smooth} have been successfully applied to neural machine translation (NMT) for domain adaptation with retrieval pipelines.
Given an advanced MT model, they generally involve two steps: 
\begin{itemize}
    \item It builds a cached memory, usually called $datastore$, in advance by extracting the context representations of the penultimate layer of the given NMT model corresponding to each target token from in-domain data.
    \item At inference, it retrieves the $k$ nearest neighbors of the context representation for each generated token from the constructed datastore and then integrates external $k$NN translation probabilities derived from these retrievals to adjust the translation.
\end{itemize}

\begin{table}[t]
    \centering
    \begin{tabular}{c|cc}
    \toprule
        Model  & speed (token/s) & BLEU \\
       \hline 
        MT              &   913.48   & 37.50 \\
        AK-MT ($k$=4)   &   642.43    & 46.32 \\
    \bottomrule
    \end{tabular}
    \caption{The inference speed comparison on the same IT-domain test set. AK-MT denotes the adaptive $k$NN-MT.}
    \label{tab:intro-inference-times}
\end{table}

The accessibility of any provided datastore during translation makes them interpretable.
Meanwhile, the reliability of these approaches gives the credit to the datastore quality.
In spite of significant translation improvements, analyses on the datastore behavior have not been fully explored yet.
We empirically observe that the construction of datastore is not optimal for retrieval from two aspects: retrieval latency and semantic distribution.

\begin{figure}[t]
    \centering
    \includegraphics[width=0.45\textwidth]{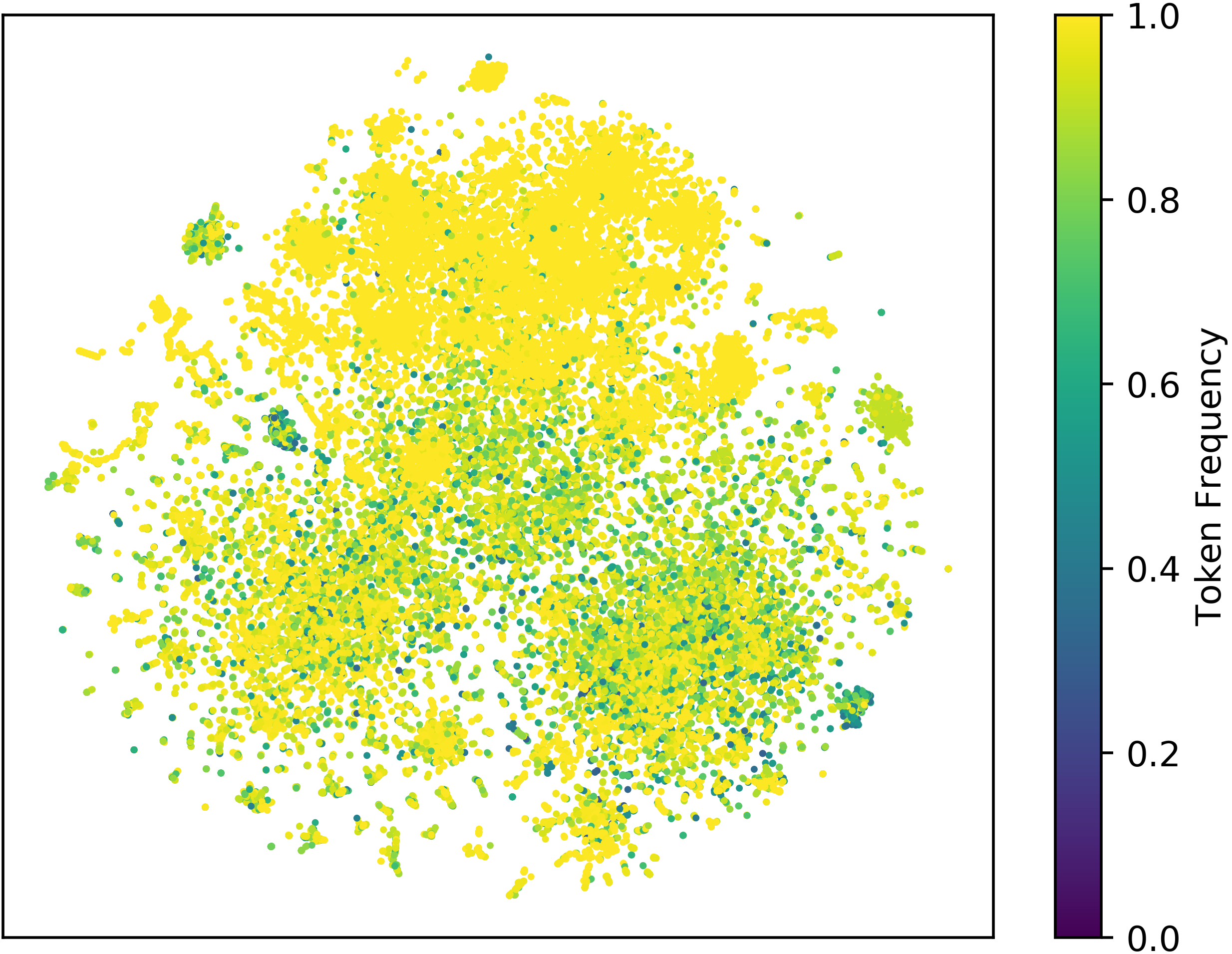}
    \caption{t-SNE visualization of IT domain features. Darker nodes denote lower frequency tokens.}
    \label{fig:visualization}
\end{figure}

\textbf{Retrieval Latency.}
As shown in Table~\ref{tab:intro-inference-times}, we compare both translation performance and speed between a pre-trained NMT model \cite{facebook} with 270M parameters and the adaptive $k$NN-MT \cite{adaptiveknnmt} system originated from the former on the same hardware (a P100-16GB GPU with 18 cores Intel Xeon Gold 6240 CPU @ 2.60GHz), where the later is the most advanced retrieval-based NMT model so far.\footnote{The speed comparison is based on the implementation released at https://github.com/zhengxxn/adaptive-knn-mt}
It indicates that the heavy computation of retrieval within a datastore causes increased latency and makes it less practical in real-time scenarios.
To address this problem, we propose an efficient pruning strategy to decrease the datastore redundancy so as to deal with the trade-off between the speed and the quality.

\textbf{Semantic Distribution.}
For robust token-to-token retrieval, tokens with similar context are expected to be distributed close to each other to form separable and compact semantic clusters, otherwise semantic noise may hurt the retrieval effectiveness.

To explore the potential of $k$-nearest retrieval, we visualize the feature distribution of a datastore built on the IT-domain corpus \cite{multi-domain-corpora} in Figure~\ref{fig:visualization}. 
For the datastore constructed in the traditional way, we have 2 important findings. 
One is that the majority tokens are distributed in the overlapped area regardless of frequency. 
The other is that even the overall distribution shows a clustering effect, only a few small clusters are correctly classified with respect to frequency.
Intuitively, these findings will directly and negatively affect the distance-based retrieval. 

Moreover, as \citep{featurecompression} suggest, the dimension is highly related to retrieval speed.
Preliminary studies on $k$NN-LM \cite{efficient-knn-lm} indicate that traditional feature reduction algorithms could only maintain the original performance until the context feature dimension is reduced to a minimum required size (e.g., for feature dimension 1024, PCA requires at least 512). 
For NMT model, it is still challenging to reduce the feature dimension to its 10\% (e.g., from 1024 to <100). 
To tackle this problem, we design a cluster-based training strategy where an external light-weight feature reduction network is learnt in a contrastive training manner to maximize the margin between context semantic clusters. 
In our experiments, we can even cut out 93.75\% of the original feature size.

In summary, our main contributions are two-fold:
\begin{itemize}
    \item We propose a cluster-based Compact Network to reduce the dimension of the semantic representations and improve the translation performance by making different tokens separable to refine the retrieval results.
    \item We further propose a cluster-based pruning strategy by filtering redundant representations in the datastore so that our proposed methods could significantly decrease the translation latency during inference.
\end{itemize}
Experiments on multi-domain machine translation benchmarks indicate that our proposed methods are superior to existing retrieval-based machine translation systems in terms of both speed and quality.

\section{Related Work and Background}
\label{sec:background}
In this section, we will briefly introduce the background of the adaptive $k$NN-MT \cite{adaptiveknnmt}.
Adaptive $k$NN-MT is derived from $k$NN-MT \cite{knnmt} by inserting a light-weight Meta-$k$ Network that fuses $k$NN retrievals with various $k$ to alleviate the possible noise induced by a single $k$.
Formally, it is formulated as two steps: target-side datastore creation and Meta-$k$ Network predictions.

\textbf{Target-side Datastore Creation.}
The datastore constists of a set of key-value pairs.
Given a bilingual sentence pair $(s, t)$ in a corpus $(S, T)$, a pre-trained general domain NMT model autoregressively extracts the context representation $h_i$ of the $i$-th target token conditioned on both source and target context $(s, t_{<i})$, denoted as $h_i = f(s, t_{<i})$.
The datastore is finally constructed by taking $h_i$ as keys and $t_i$ as values:
\begin{equation*}
(\mathcal{K}, \mathcal{V}) = \bigcup_{(s, t) \in (\mathcal{S}, \mathcal{T})} \{(h_i, t_i), \forall ~t_i \in t \}. 
\end{equation*}

\begin{figure*}[t]
    \centering
    \includegraphics[width=0.82\textwidth]{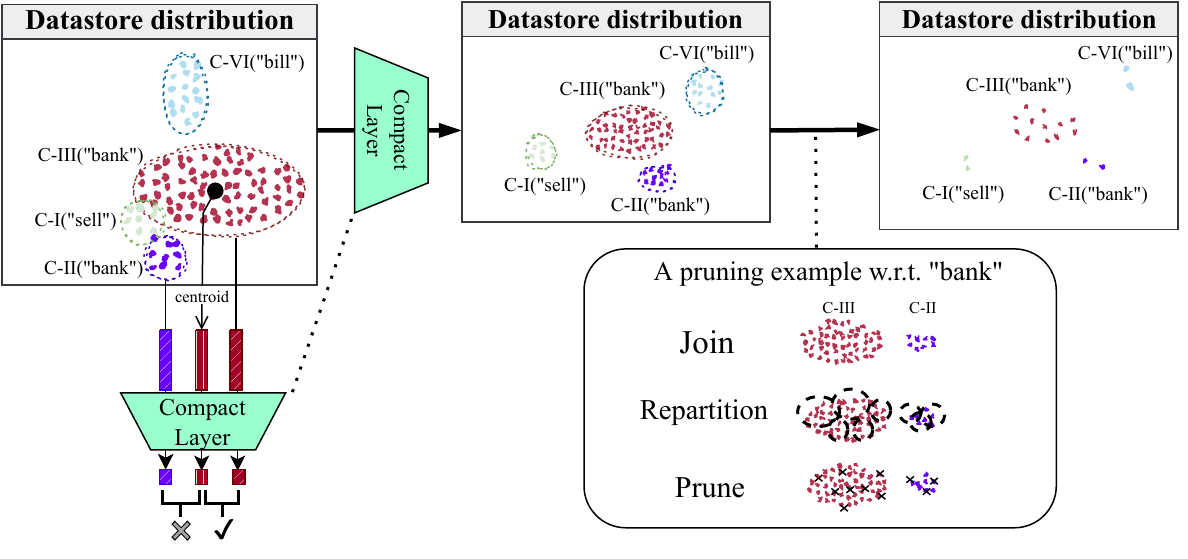}
    \caption{The diagram of the proposed approach. C-*("\#") denotes the *th cluster of token "\#". First, the cluster-based Compact Network is used to reduce the key's dimensionality of the original datastore and a new datastore is reconstructed. Then the cluster-based pruning is applied to reduce the datastore size.}
    \label{fig:our-approach}
\end{figure*}

\textbf{Meta-$k$ Network Prediction.}
Meta-$k$ Network ($f_{\beta}$) is a two-layer feed-forward network followed by a non-linear activation function.
Based on the constructed datastore, it considers a set of different $k$s that are smaller than an upper bound $K$.
The standard setting for $k$ is $\mathcal{Q} = \{ 0 \} \cup \{ k_r \in \mathbb{N} \mid \log_2 k_r \in \mathbb{N}, k_r \leq K \}$.
$K$ nearest neighbors of the current context query $\hat{h}_i$ from the datastore are first retrieved at the $i$-th decoding step.
Then the square of $l_2$ distance from $\hat{h}_i$ to each neighbor $(h_j, v_j)$ is denoted as $d_j=\Vert h_j, \hat{h}_i \Vert^2$.
And the number of distinct values in top $j$ neighbors are denoted as $c_j$.
The normalized weights of each available $k$ are computed as:
\begin{equation*}
    p_{\beta}(k) = \textrm{softmax} (f_{\beta}([d_1,...,d_K;c_1,...,c_K]))
\end{equation*}
where $f_\beta$ denotes the Meta-$k$ Network.
For $k_r \in \mathcal{Q}$, the word prediction  probability over the vocabulary w.r.t each neighbor is computed via the Gaussian kernal function:
\begin{align*}
    p_{k_r\textrm{NN}}(y_i | x, \hat{y}_{<i})
    & \propto \\
    \sum_{\{(h_j, v_j)~|~j \leq k_r,j\in \mathbb{N}\}}
    & \mathbb{1}_{y_i = v_j} \exp (\frac{-\Vert h_j, \hat{h}_i\Vert^2}{T})
\end{align*}
where $T$ denotes the temperature hyper-parameter.
The ultimate prediction probability is a weighted ensemble:
\begin{equation*}
p(t_i| s,\hat{t}_{<i}) =  \sum_{k_r \in \mathcal{Q}} p_{\beta}(k_r) \cdot p_{k_r\textrm{NN}}(t_i|s, \hat{t}_{<i})
\end{equation*}

Note that a validation set is usually required to study the Meta-$k$ Network before predicting on test sets. During training, only the parameters of the Meta-$k$ Network need to update.

\section{Our Approach}
\label{sec:our-approach}
As shown in Figure~\ref{fig:our-approach}, our proposed approach focuses on datastore reconstruction from the perspectives of feature compression and size pruning by utilizing cluster-based signals.

\subsection{Cluster-Based Feature Compression}

From Figure~\ref{fig:visualization}, we observe that spatially close context representations may have noisy and different semantics.
During inference, it may lead to unreliable neighbors for retrieval-based NMT (see examples in Appendix \ref{sec:case-analysis} ``Case Analysis'') due to the entanglements from these noisy context space. 
We hypothesize that the reasons may be three-fold. 
First, the pre-trained NMT model on general domain lacks target domain-specific knowledge. 
Second, the high dimensional semantic space is too sparse and may contain some noisy underlying components. 
Third, the likelihood-maximization objective from the logits by dot-production enforces the alignment of vector directions, which is inconsistent with the spatially close expectation for the sake of both direction and length.

To address these issues, we propose a one-plus-one ($f_{\alpha}$+$f_{\theta}$) Compact Network on top of the pre-trained NMT model. 
The first ``one'' module is to transform the coarse-grained semantics of the pre-trained NMT into the fine-grained semantic clusters. 
The second ``one'' module is used to calculate our designed loss function.

\begin{figure}[t]
    \centering
    \includegraphics[width=0.48\textwidth]{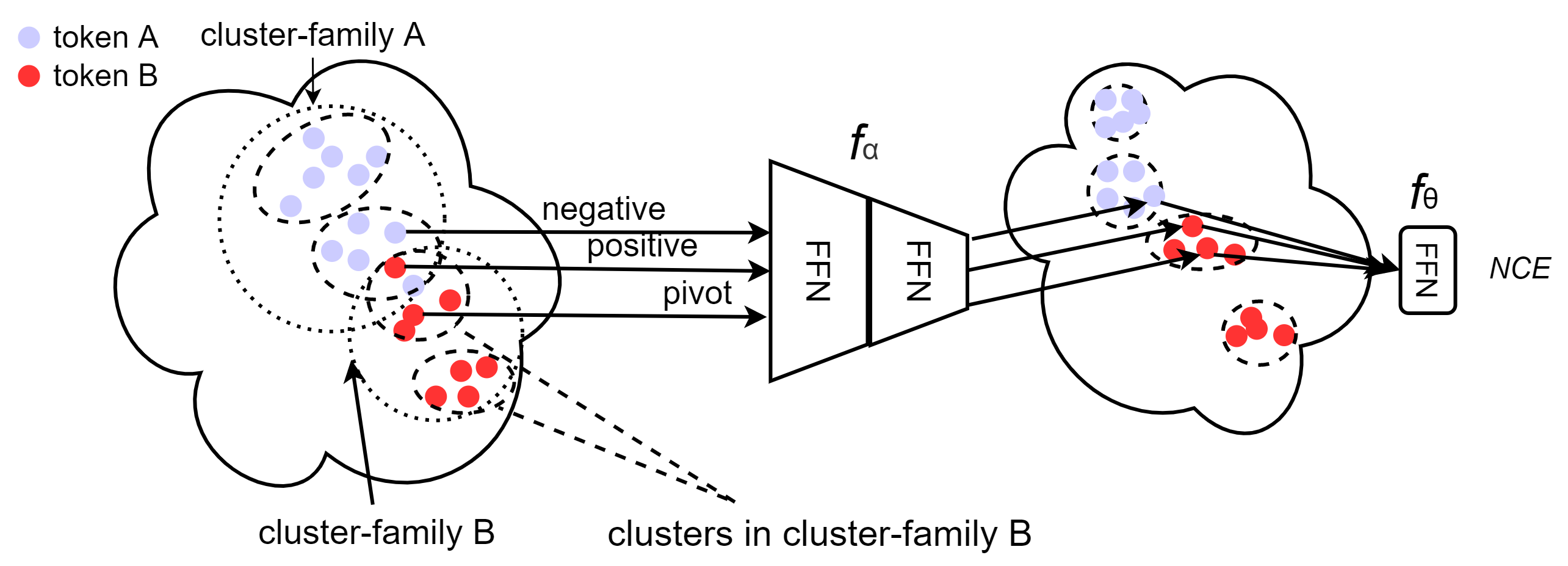}
    \caption{The Compact Network illustration. $f_\alpha$ is for dimension reduction and $f_\theta$ is for NCE training.}
    \label{fig:compact-network}
\end{figure}

To obtain coarse-grained semantic clusters, we first follow the  method described in ``Target-side Datastore Creation'' of Section \ref{sec:background} to create the in-domain datastore. 
For context representations (keys) with the same target token (value), we conduct target-side clustering for the representations, shown as the left clusters in Figure~\ref{fig:compact-network}. 
We denote the resulted clusters from the same value as the \textbf{cluster family} for the corresponding target token. 
Due to the distance-based clustering, it is guaranteed that clusters within each cluster family are not overlapped at all. 
However, different cluster families will have large overlapped space according to Figure~\ref{fig:visualization}. 
Therefore, our main purpose is to construct a transform that can make the cluster families separable as well. 

The proposed light-weight Compact Network in Figure~\ref{fig:compact-network} is desired to fulfill above purpose and compress the feature dimension. 
The first two-layer perceptron is applied for representation compression:
$
f_{\alpha}(\cdot) = \rm FFN_2(\sigma~(\textit{\rm FFN}_1(\cdot)))
$,
where $\sigma(\cdot)$ denotes the Sigmoid function.
The last layer $f_\theta$ is attached for transferring the compressed representations into classification logits where the output dimension depends on the number of designed categories. 
Note that the $f_\theta$ layer is discarded at inference.

In order to obtain the separable cluster families after $f_\alpha$, we are motivated to consider several candidate contrastive regularizations to train the Compact Network.

\textbf{Triplet Noise-Contrastive Estimation (NCE).}
For each cluster in one particular cluster family, two semantic representations are randomly sampled, one as the pivot example $v_{*}$ and the other as the positive example $v_{+}$. 
From the cluster in a different cluster family, another semantic representation is randomly selected as the negative example $v_{-}$. 
Then we conduct NCE \cite{nce} with binary classification on ~\{pivot, positive\} and \{pivot, negative\} to predict which pair belongs to the same cluster.
\begin{equation*}\begin{aligned}
    \min_{f_{\theta},f_{\alpha}} & -\log(\sigma (f_{\theta}([f_{\alpha}(v_+); f_{\alpha}(v_{*})]))) \\
    &-\log(1-\sigma (f_{\theta}([f_{\alpha}(v_-); f_{\alpha}(v_{*})])))
\end{aligned}\end{equation*}
where the output dimension of $f_{\theta}$ is 1.

\textbf{Triplet Distance Ranking.}
This is similar to the Triplet NCE. The differences are that (1) we remove the $f_\theta$ layer and (2) the objective is modified as a ranking loss by minimizing the $l_2$ distance between the pivot and positive examples as well as maximizing the distance between the pivot and negative ones:
\begin{align*}
    \min_{f_{\theta},f_{\alpha}} \Vert & f_{\alpha}(v_+) - f_{\alpha}(v_{*})\Vert_2 \\
    &+ 1 / \Vert f_{\alpha}(v_-)-f_{\alpha}(v_{*}) \Vert_2
\end{align*}

\textbf{Word Prediction Loss.} To compensate the loss of linguistic information that NCE may ignore, the traditional word prediction NMT loss is also used to train the Compact Network. 
In this scenario, the output dimension of $f_\theta$ is the vocabulary size of the corresponding target language.

In addition, we find that dynamic pivot selection leads to unstable training as the compressed representations are forced to update toward various directions. 
For each cluster, we modify the dynamic pivot as a static pivot, by fixing it as the centroid. 
After the training converges, we can construct a new feature-compressed datastore with the output of $f_\alpha$, which is used for query retrieval during the $k$NN-MT inference.

\subsection{Cluster-Based Pruning}

Apart from feature reduction, the number of key-value pairs in the compressed datastore is crucial for the translation latency as well, hence redundant tokens are encouraged to be pruned.
In literature, phrase-level pruning strategies have proved efficient for statistical machine translation (SMT) \cite{entropy-smt-prune, system-smt-prune}. 
Each record in the phrase table reflects a similar semantic unit, hence one could prune parts of the records that share similar statistics, e.g., translation quality, translation cost, etc. 

\begin{figure}[t]
    \centering
    \includegraphics[width=0.48\textwidth]{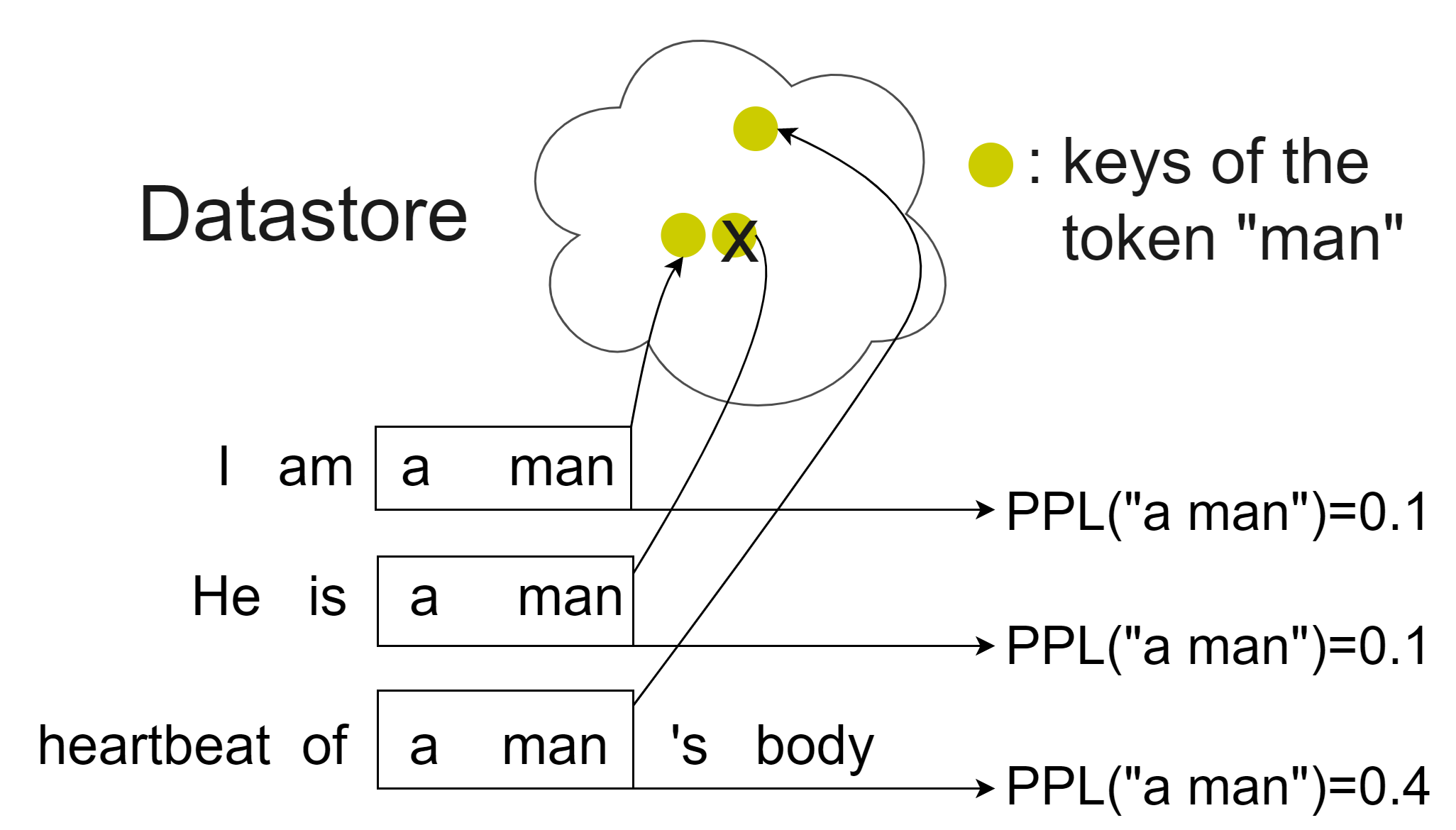}
    \caption{An example of redundant bigram "a man" with similar translation costs. "X" denotes that the node with similar PPL will be randomly deleted in pruning.}
    \label{fig:example-translation-cost}
\end{figure}

Enlightened by SMT, we propose an efficient pruning strategy based on $n$-gram metrics on the original semantic representation space. 
Intuitively, the entry of a key-value pair in the datastore is \textbf{redundant} if there are other key-value pairs (with the same value) holding for that the difference of their perplexity (PPL) values is smaller than a given threshold $\epsilon$ (an example is represented in Figure~\ref{fig:example-translation-cost}).

To make it concrete, we decrible the \textbf{translation cost} as follows. 
For a given $n$-gram phrase ($t_{i-n+1}, t_{i-n+2}, ..., t_{i}$) in the translation with the corresponding token-level translation probability $p(t_{j}|s, t_{<{j}})~\forall j \in \{i, i-1, ..., i-n+1\}$, we measure the translation cost of its last token (desired value in datastore) as the perplexity (PPL) of the $n$-gram phrase.
However, when $n$ is fixed, $n$-gram phrases are not always meaningful because some translations are independent of its previous target-side context \cite{entropy-smt-prune}.
Hence we do not directly adopt the naive PPL as a stable translation cost but truncate it in a heuristic way. 
We search for the minimal PPL among all consecutive sub-sequences ending with that last token. 
Formally, given a bilingual sentence pair $(s,t)$, we define the translation cost for each target token $t_i$:
\begin{align*}
    c_{t_i} = \min_{b \in \{1, 2, ..., n\}} & {\rm PPL} (
    p(t_{i-b+1}|s, t_{<{i-b+1}}), ~..., \\
    & ~p(t_{i-1}|s, t_{<i-1}), ~p(t_i|s, t_{<i}))
\end{align*}

Then we can add the translation cost into the feature-compressed datastore.
\begin{align*}
 ((\mathcal{K}, & \mathcal{C}), \mathcal{V}) = \\
 &\bigcup_{(s, t) \in (\mathcal{S}, \mathcal{T})} \{((f_\alpha(h_i), c_{t_i}), t_i), \forall ~t_i \in t \}
\end{align*}

\begin{table}[t]
    \centering
    \begin{tabular}{l}
      \toprule
      \textbf{Algorithm 1 Cluster-Based Pruning}\\
      \hline
      \textbf{Input}:\\
      ~~~~~The expected pruning rate $r$. \\
      ~~~~~The translation cost threshold $\epsilon$. \\
      ~~~~~A preprocessed datastore (($\mathcal{K}, \mathcal{C} $), $ \mathcal{V}$). \\
      \textbf{Output}:\\
      ~~~~~A new pruned datastore ($\mathcal{K}_{new}, \mathcal{V}_{new}$). \\
      1. Greedy Clustering On Translation Costs. \\
      ~~~~~$G \gets \emptyset$. \\
      ~~~~~\textbf{For each} $v_i$ in set($\mathcal{V}$) \textbf{do} \\
      ~~~~~~~~~get collection ($\mathcal{K}_{v_i}$, $\mathcal{C}_{v_i}$) paired with $v_i$ \\
      ~~~~~~~~~split$_{\mathcal{C}_{v_i}} \gets$ ~ cluster($\mathcal{C}_{v_i}$, $\epsilon$) \\
      ~~~~~~~~~$\mathcal{K}_{\rm split} ~~ \gets$ ~ map(split$_{\mathcal{C}_{v_i}},\mathcal{K}_{v_i}$) \\
      ~~~~~~~~~$G$.extend(~zip($\mathcal{K}_{\rm split}, \mathcal{V}_{v_i}$)~) \\
      2. Uniform Pruning. \\
      ~~~~~$D_{\rm{new}} \gets \{\}$. \\
      ~~~~~\textbf{For each} ($k, v$) in $G$ \textbf{do} \\
      ~~~~~~~~~~$k_*, v_*$ =  sample\_by\_rate(($k, v$), $r$) \\
      ~~~~~~~~~~$D_{\rm new}$.update($k_*, v_*$) \\
      ~~~~~return $D_{\rm{new}}$ \\
      \bottomrule
     \end{tabular}
     \label{table:algorithm}
 \end{table}

For the augmented datastore described above, we only apply propagation-based clustering \cite{dbscan, birch} upon the translation cost $c_{t_i}$ to get cost-similar groups, and partition the semantic representations in accordance to these groups. 
To get pruned datastore, we adopt uniform sampling on each group and collect them into a small key-value paired datastore. 
This algorithm is summarized in Algorithm~1. 

In brief, our efficient cluster-based $k$-nearest neighbor machine translation can be concluded into the following steps.
\begin{itemize}
    \item We adopt the original datastore to train Compact Network while the parameters of NMT are frozen.
    \item  We adopt the validation set to train the Meta-$k$ Network while the parameters of NMT and Compact Network are fixed.
    \item We reconstruct the feature-compressed datastore and prune it into a small datastore using our proposed $n$-gram pruning algorithm that will be eventually used for testing. 
\end{itemize}

\begin{table*}[t]
    \centering
    \begin{tabular}{c|ccccc|ccccc}
      \toprule
      \textbf{Dataset} &
      \multicolumn{5}{c}{\textbf{Statistics of training sets}} &
      \multicolumn{5}{c}{\textbf{Statistics of test sets}} \\
      \hline
      Domain &
      \textbf{Koran} & \textbf{IT} & \textbf{Medical} & \textbf{Law} & \textbf{Sub} &
      \textbf{Koran} & \textbf{IT} & \textbf{Medical} & \textbf{Law} & \textbf{Sub} \\
      sentence & 222K & 248K & 18K & 467K & 12.4M & 2K & 2K & 2K & 2K & 2K \\
      token  & 0.5M & 3.6M & 6.9M & 19M & 154M & 58K & 34K & 57K & 81K & 25K \\
      \bottomrule
    \end{tabular}
    \caption{The statistics of datasets in all experiments. ``Sub" denotes Subtitles.}
    \label{tab:datasets}
\end{table*}

\section{Experiments}

We carried out a series of experiments to evaluate the proposed non-parametric NMT against the previous advanced counterpart on several translation benchmarks. 

\subsection{Datasets}

We followed \citep{adaptiveknnmt} to conduct all experiments on five widely used machine translation benchmarks of unique domains, including IT, Koran, Medical, Law and Subtitles.
The first four domains were also used in \citep{adaptiveknnmt} while the last Subtitles dataset contains a large number of target tokens, which is hence suitable to explore our pruning strategy.
The statistics of these datasets are shown in Table~\ref{tab:datasets}.
We tokenized sentences using Moses\footnote{https://github.com/moses-smt/
mosesdecoder} and split words into subword units \citep{bpe} with the bpe-codes provided by \cite{facebook}.
We applied the product quantizer with the inverted file system based on Faiss\footnote{https://github.com/facebookresearch/faiss/} to quantize the datastores and conduct retrieval. 
The hyper-parameters of Faiss are provided in Appendix~\ref{appendix:hyper-parameter}.

\subsection{Clustering Algorithm Selection}
The determination of clustering algorithms depends on computation complexity and clustering effectiveness.
\begin{itemize}
    \item As semantic clusters in a large datastore are vague and it is hard to determine the prior quantity of clusters existing in a large datastore, clustering algorithms that hold a static cluster quantity in advance (e.g., $k$-Means \cite{k-means}) are not fit for dataset partitioning.
    \item Besides, clustering complexity is not tolerant in practice when it increases up to $O(N^2)$ (e.g., Affinity Propagation \cite{ap}) since $N$ is usually extremely large for a high-quality datastore.
\end{itemize}
We eventually chose two classical clustering algorithms from candidates for exploration in our experiments: DBSCAN \cite{dbscan} and Birch \cite{birch}.
DBSCAN was applied for clustering datastore with 100M- nodes while BIRCH was applied for clustering datastore with 100M+ nodes for the sake of computation-and-quality trade-off.
In our experiments, We adopted the scikit-learn clustering implements.\footnote{https://scikit-learn.org/stable/modules/clustering.html}

\subsection{Baselines}

We adopted the following models as our baselines.
\begin{itemize}
    \item \textbf{Base NMT.} This is the winner model \cite{transformer} of WMT'19 German-English News translation task\footnote{http://www.statmt.org/wmt19/} provided by \citep{facebook}, which is also used in \cite{adaptiveknnmt}. It is a Transformer model \cite{transformer} with hidden size 1024.
    \item \textbf{Adaptive $k$NN-MT} \cite{adaptiveknnmt}. This is the benchmark model of our work.
\end{itemize}

In our modifications, as expected to reduce the dimension to <10\% of its original size, we did greedy searching in [16, 32, 64, 128] to obtain the optimal 64 as $f_\alpha$'s output dimension on the IT domain validation set and then used this setting in all experiments. 
The detailed dimension related analysis can be found in Appendix~\ref{sec:compact-dimension-decision}. 
Similarly we used grid search and selected bigram in the clustering-based pruning algorithm.

\subsection{Evaluation}
\begin{table}[t]
    \centering
    \begin{tabular}{l|c}
        \toprule
        Model  & BLEU \\
        \hline
        NMT                       & 38.35 \\
        adaptive $k$NN-MT         & 47.20 \\
        ~~~~+feature-wise PCA     & 46.84 \\
        ~~~~+weight-wise SVD      & 45.96 \\
        $[$DY$]$ CKMT+DR          & 37.10 \\
        $[$DY$]$ CKMT+WP          & 46.41 \\
        $[$DY$]$ CKMT+NCE         & 46.58 \\
        $[$DY$]$ CKMT+NCE+DR      & 37.33 \\
        $[$DY$]$ CKMT+NCE+WP      & 46.42 \\
        $[$DY$]$ CKMT+NCE+CL      & 47.48 \\
        $[$ST$]$ ~CKMT+NCE+CL     & \textbf{47.94} \\
        $[$ST$]$ ~CKMT+NCE+CL+DR  & 47.64 \\
        $[$ST$]$ ~CKMT+NCE+CL+WP  & 46.88 \\
    \bottomrule
    \end{tabular}
    \caption{The BLEU performance comparison of the feature reduction methods on the IT domain. All retrieval $k$ is set to 4.
    DR, NCE and WP denote the distance ranking, noise-contrastive estimation and word prediction objectives, respectively. CL denotes that all the tokens are clustered and then the triplets are selected based on these clusters. [DY] denotes that the pivot is dynamically selected while [ST] denotes static pivot selection.}
    \label{tab:results-compact-network}
\end{table}

\begin{table}[t]
    \centering
    \setlength{\tabcolsep}{1.5pt}
    \begin{tabular}{c|cccc|c}
    \toprule
    \multirow{2}{*}{Model}  & \multicolumn{4}{c|}{Tested Domain} & \multirow{2}{*}{Avg.} \\
       \cline{2-5}
                            & IT    & Koran & Law    & Medical   & \\
        \hline
        NMT                 & 38.35 & 16.26 & 45.48  & 39.99     & 35.02 \\
        adaptive $k$NN-MT   & 47.20 & 19.39 & 62.64  & 55.71     & 46.24 \\
        CKMT*               & \textbf{47.94} & \textbf{19.92} & \textbf{62.98}  & \textbf{56.92} & \textbf{46.94} \\
        G-CKMT*       & 47.27 & 19.84 & 62.55 & 56.52 & 46.55 \\
    \bottomrule
    \end{tabular}
    \caption{The translation BLEU comparison in different domains. G-CKMT* denotes that the Compact Network of CKMT* was trained using the general Wikimatrix datastore.}
    \label{tab:results-compact-network-4-domains}
\end{table}

All experiments were conducted on a P100-16GB GPU with 18 cores Intel(R) Xeon(R) Gold 6240 CPU @ 2.60GHz except for the experiments in Subsection~\ref{sec:rexperiment-pruning-experiments} where we used 2 GPU cards to load a larger datastore. 
All translation results were evaluated in case-sensitive detokenized BLEU with SacreBLEU \cite{sacrebleu}. 

\subsection{Results}


For simplicity, we refer to the base NMT model equipped with the proposed Compact Network as \textbf{CKMT} and further equipped with the pruned datastore as \textbf{PCKMT} in this section.

\subsubsection{Performance of the Compact Network}

On the IT domain, we first evaluated the compact layer settings mentioned in Section~\ref{sec:our-approach}, as well as two traditional feature reduction algorithms: Principal Component Analysis (PCA) used in \cite{efficient-knn-lm} and Singular Value Decomposition (SVD).
We applied the PCA solution to learn feature-wise linear projection while the SVD solution to learn matrix-wise projection that decomposes the weight ($W$) of the last layer of the base NMT model into three matrices:
\begin{equation*}
    W_{1024*vocab\_size} = S_{1024*64}U_{64*64}V_{64*vocab\_size}
\end{equation*}
Then $f_\alpha$ can be replaced  by an FFN layer with the weight $S_{1024*64}U_{64*64}$ but without bias.

As shown in Table~\ref{tab:results-compact-network}, the best CKMT solution is equipped with the Compact Network trained using NCE+CL+DR. It outperforms the adaptive $k$NN-MT by 0.74 BLEU.
Being consistent with \cite{efficient-knn-lm}, we find that it is difficult to use the 1024-to-64 feature-wise PCA to maintain the translation performance with such a low dimension.
Basically, the distance ranking loss causes serious performance degradation. 
We assume that the distance minimization restraint is too strict to optimize a small datastore since both the direction and the length of a semantic vector have already been optimized.
Though the word prediction (WP) can recover semantic information, its $f_\theta$ has too many parameters to be optimized on the limited IT domain datastet compared with NCE alone. 
Besides, we attribute the improvement obtained by the clustering (CL) to the introduced semantic disambiguation. 
Finally, the static pivot selection (ST) achieves an improvement of 0.46 BLEU against the dynamic method.

We refer to the best setting [ST] CKMT+NCE+CL as CKMT*, and report the results against the adaptive $k$NN-MT on various domains in Table~\ref{tab:results-compact-network-4-domains}.
CKMT* gains an average improvement of 0.70 BLEU over the adaptive $k$NN-MT which indicates that our proposed Compact Network refines the retrieval for machine translation.

\begin{table}[t]
    \centering
    \begin{tabular}{ccc}
    \toprule
        Rate  & Datastore Size & BLEU\\
        \hline
        100\% & 3.6M & 47.94\\
        80\%  & 2.9M & 47.67 \\
        60\%  & 2.2M & 47.57 \\
        40\%  & 1.4M & 47.29 \\
        20\%  & 0.7M & 46.98 \\
        1\%  & 0.04M & 46.21 \\
    \bottomrule
    \end{tabular}
    \caption{Performance of CKMT* using decreasing rates of data to train the Compact Network at state I.}
    \label{tab:results-limited-data-experiments}
\end{table}

\textbf{The Compact Network Training with Limited Data.}
It is unclear how much data are adequate at training-stage I.
Hence, we gradually reduce the number of key-value pairs in the datastore to train the Compact Network as shown in Table~\ref{tab:results-limited-data-experiments}. 
As the number decreases, the performance degrades slowly. 
When we use only 40\% of the datastore for training, CKMT still outperforms the adaptive $k$NN-MT. 
It indicates that our proposed Compact Network is efficient and requires a small amount of key-value pairs to compress the semantic representations with contrastive loss.

\textbf{Cross Domain Generalization.}
Is there a general Compact Network that is capable to generalize to different domains?
If so, we will save the cost to train an unique Compact Network for various target domains.
To explore this, we trained the Compact Network in a general domain with the large-scale Wikimatrix Corpus \cite{wikimatrix} and evaluated its behavior on various target domains.
As the last row of Table~\ref{tab:results-compact-network-4-domains} shows, it is interesting that the general CKMT* drops only 0.39 BLEU compared with 4 domain-specific datastores, and it still outperforms the adaptive $k$NN-MT by 0.31 BLEU. 
Overall speaking, the Compact Network generalizes well across different domains.

\subsubsection{Performance of Pruning Methods}
\label{sec:rexperiment-pruning-experiments}

We tested our language-wise PPL-based pruning methods with several pruning strategies as follows.
\begin{itemize}
    \item \textbf{Spatially Pruning by Distance (SP)}. It is a naive pruning strategy using distance-wise solution by cutting off nodes with low probability according to the distance from each node to its cluster center.
    \item \textbf{Low Translation Probability Pruning (LTP)}. Tokens translated with low probabilities tend to have poor translation quality, and will be pruned for datastore stability.
    \item \textbf{High Translation Probability Pruning (HTP)}. As the $k$NN probabilities are beneficial for hart-to-translate words that NMT cannot handle, it would be more encouraged to restore the tokens wrongly translated by the base NMT. In this sense, tokens paired with high confidence will be pruned.
    \item \textbf{Random Pruning (RP)}. We also perform the random pruning strategy alone for the target-side clusters, as the step 2 introduced in Algorithm 1.
\end{itemize}

\begin{table}[t]
    \centering
    \setlength{\tabcolsep}{1.5pt}
    \begin{tabular}{l|cccc|c}
    \toprule
    \multirow{2}{*}{Model} & 
    \multicolumn{4}{c|}{Domain} &
    \multirow{2}{*}{Avg.} \\
    \cline{2-5}
    & IT & Koran & Law & Medical\\
     \hline
     CKMT*                & 47.94 & 19.92 & 62.98 & 56.92 & 46.94 \\
     \hline
     CKMT*+SP             & 43.01 & 19.50  & 59.40 & 52.16 & 43.52\\
     CKMT*+LTP            & 46.78 & 19.28 & \textbf{61.96} & 55.21 & 45.81 \\
     CKMT*+HTP            & 45.95 & \textbf{20.10} & 59.51 & 55.14 & 45.18 \\
     CKMT*+RP             & 46.38 & 19.99 & \textbf{61.96} & \textbf{55.45} & 45.85 \\
     CKMT*+Ours           & \textbf{47.06} & 20.01 & 61.72 & 55.33 & \textbf{46.03} \\
    \bottomrule
    \end{tabular}
    \caption{Translation BLEU Results on 4 different domains with 10\% pruning rate. $k$ was set to 4. Note that CKMT* in the first row used the full datastore.}
    \label{tab:results-small-domain-pruning}
\end{table}  

The results on 4 different domains are shown in Table~\ref{tab:results-small-domain-pruning}.
Since the datastore size remains the same (10\% pruned) for all pruning methods in Table~\ref{tab:results-small-domain-pruning}, there is no much retrieval speed difference among these methods. 
Our cluster based pruning strategy generally achieves the smallest degradation. 
Though other strategies obtain impressive\footnote{This is in comparison to previous studies (e.g., \cite{efficient-knn-lm})
that usually fail to maintain model performance when datastores are pruned to a large extent.} results on a few domains (e.g., 10\% pruned CKMT*+HTP outperforms non-pruned CKMT* by 0.18 BLEU on the Koran test set) since previous studies (i.e, \cite{efficient-knn-lm}) our cluster-based pruning strategy performs the most stably on average. 
Note that the random pruning strategy is simple yet effective, which coincides with \cite{efficient-knn-lm}.

\begin{figure}[t]
    \centering
    \includegraphics[width=0.48\textwidth]{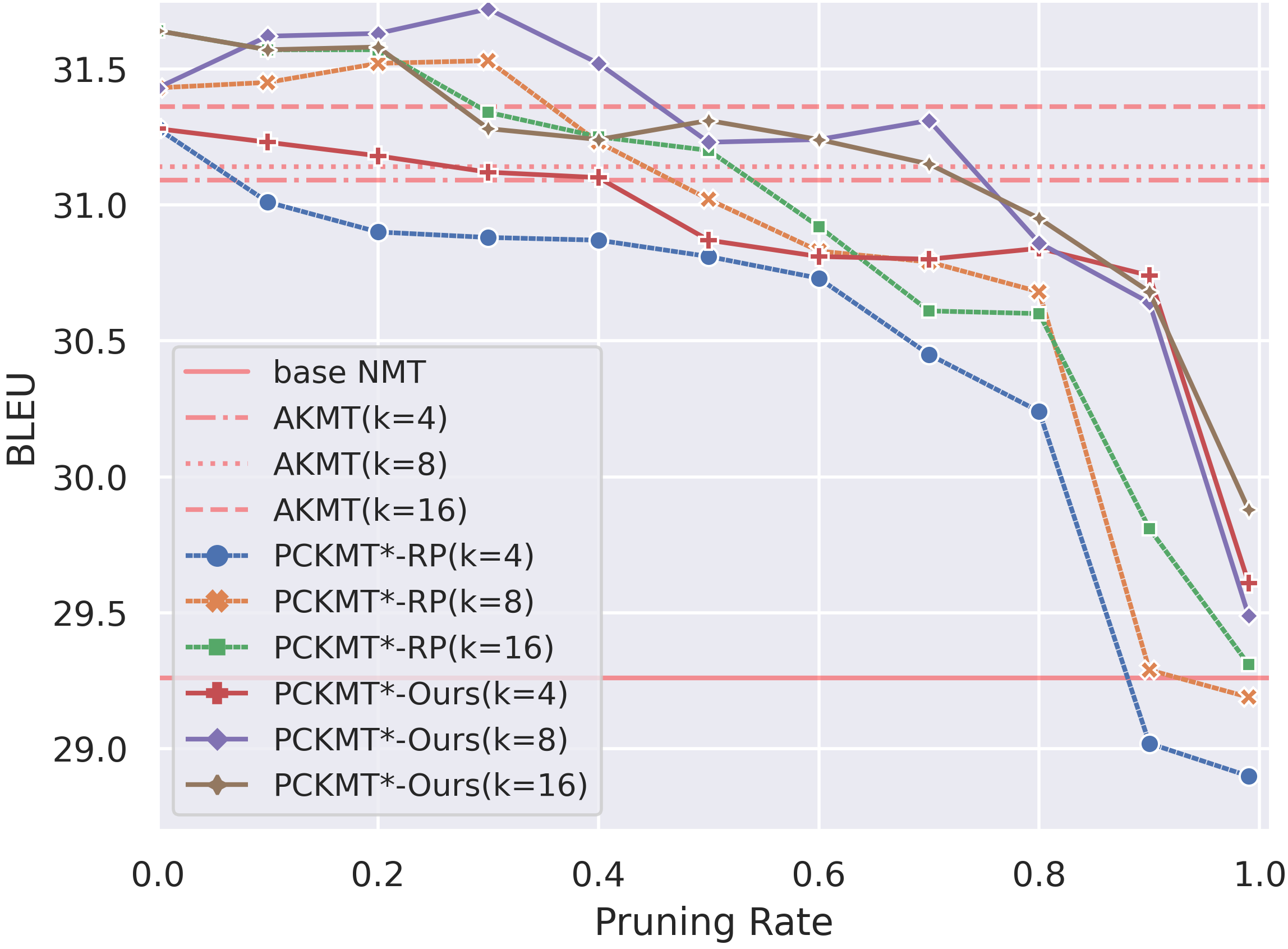}
    \caption{The BLEU comparison of pruning experiments on the Subtitles domain with increasing pruning rates. AKMT denotes the non-pruned adaptive $k$NN-MT.} 
    \label{fig:results-subtitile-pruning}
\end{figure}

\begin{table*}[t]
    \centering
    \small
    \begin{tabular}{c|c|c|cc|cc}
    \toprule
    
     \multirow{2}{*}{Model}
     & \multirow{2}{*}{Batch Size}
     & \multirow{2}{*}{$k$}
     &  \multicolumn{2}{c|}{Speed}
     &  \multicolumn{2}{c}{Datastore Storage} \\
     \cline{4-7}
     & & & sentences/s & tokens/s & original & quantization\\
    \hline
    
    \multirow{2}{*}{base NMT}
    & 8  & \multirow{2}{*}{-} & 49  & 572  & \multirow{2}{*}{-} & \multirow{2}{*}{-}\\
    & 64 &                    & 166 & 1959 & &\\
    
    \hline
    \multirow{6}{*}{adaptive $k$NN-MT NMT}
    & 8  & \multirow{2}{*}{16}  & 26 (-23) & 292 (-280) & \multirow{6}{*}{295.2GB} & \multirow{6}{*}{12.2GB} \\
    & 64 &                      & 55 (-111) & 618 (-1341) & & \\
    \cline{2-5}
    & 8  & \multirow{2}{*}{8}   & 26 (-23)  & 295 (-277) & & \\
    & 64 &                      & 59 (-107) & 663 (-1296) & & \\
    \cline{2-5}
    & 8  & \multirow{2}{*}{4}   & 26 (-23)  & 296 (-276) & & \\
    & 64 &                      & 58 (-108) & 660 (-1299) & & \\

    \hline
    \multirow{2}{*}{PCKMT*-20\%}
    & 8  & \multirow{2}{*}{16}  & 34 (-15) & 384 (-188) & \multirow{2}{*}{15.9GB} & \multirow{2}{*}{9.1GB} \\
    & 64 &                      & 85 (-81) & 963 (-996) &  & \\
    \hline
    \multirow{2}{*}{PCKMT*-70\%}
    & 8  & \multirow{2}{*}{8}   & 37 (-12) & 419 (-153) & \multirow{2}{*}{6.5GB} & \multirow{2}{*}{3.98GB} \\
    & 64 &                      & 100 (-66) & 1132 (-827) &  & \\
    \hline
    \multirow{2}{*}{PCKMT*-40\%}
    & 8  & \multirow{2}{*}{4}   & 39 (-10) & 444 (-128) &  \multirow{2}{*}{12.7GB} & \multirow{2}{*}{7.0GB} \\
    & 64 &                      & 98 (-68) & 1108 (-851) & & \\

    \bottomrule
    \end{tabular}
    \caption{The computation cost of PCKMT* with no BLEU degradation compared with the adaptive $k$NN-MT. PCKMT*-\#\% denotes PCKMT* equipped with the  \#\% pruned datastore for retrieval during inference. The value in parentheses is the speed latency between the corresponding model and the base NMT.}
    \label{tab:results-tolerant-0bleu}
\end{table*}

\begin{table*}[t]
    \centering
    \small
    \begin{tabular}{cc|ccccc}
    \toprule
        \multicolumn{2}{c|}{Model}  & BLEU & Sentences/s & Tokens/s & Datastore size & Pruning rate\\
       \hline
        \multicolumn{2}{c|}{adaptive $k$NN-MT}
        &  31.36           & 58  & 660 & 154M & 0\%\\
        \multirow{2}{*}{k=16}
        & CKMT*  &  31.64           & 74  & 849  & 154M & 0\%\\
        & PCKMT* &  31.58           & 85  & 963  & 123M & 20\%\\
        \multirow{2}{*}{k=8}
        & CKMT*  &  31.43           & 78  & 890  & 154M & 0\% \\
        & PCKMT* &  \textbf{31.72}  & \textbf{91}& \textbf{1024}  & 108M & 30\% \\
        \multirow{2}{*}{k=4}
        & CKMT*  &  31.28           & 79  & 899  & 154M & 0\% \\
        & PCKMT* &  31.23           & 85  & 968  & 138M & 10\%\\
    \bottomrule
    \end{tabular}
    \caption{The optimal performances of our approach on the Subtitles test set. The batch size was fixed as 64.}
    \label{tab:results-optimal-pruning}
\end{table*}

However, we find that the in-domain data of the tested domains have limited redundancy since the average frequency of bigrams is too low (e.g., more than 0.4M unique bigrams were collected from the 3.6M IT domain datastore, on average each bigrams only has no more than 9 occurrences in the datastore).
Therefore, even 10\% pruning rate can lead to about 1 BLEU loss in Table~\ref{tab:results-small-domain-pruning}.
We leave reducing the datastore with low $n$-gram redundancy to our future work.

To further explore the potential of the pruning methods on large datastore, we conducted pruning experiments on Subtitles domain containing 154M keys.
We tested the random pruning strategy as well because it is the second competitive pruning strategy. 
As Figure~\ref{fig:results-subtitile-pruning} illustrates, the proposed PCKMT*+Ours with pruning rate 30\%  can even outperform non-pruned CKMT*. 
As the pruning rate increases, PCKMT*+Ours generally outperforms PCMKT*+RP for the same $k$. 
The performance of PCKMT*+RP drops seriously (more than 1 BLEU point) when the pruning rate $\geq$ 50\%, but PCKMT*+Ours sees a clear drop until the pruning rate $\geq$ 70\%.
When the pruning rate increases to 80+\%, PCKMT*+RP even performs worse than the base NMT, but PCKMT*+Ours still outperforms it by a large margin.
These results suggest that the proposed cluster-based pruning algorithm is effective for datastore reduction.

In Table~\ref{tab:results-tolerant-0bleu}, we further evaluated the computation cost of CKMT* with the same BLEU performance as the adaptive $k$NN-MT. 
With the same $k$ and the batch size, PCKMT* achieves 27\%\textasciitilde57\% less speed latency compared with the adaptive $k$NN-MT.
In addition, we compared our optimally performed model with baselines in Table~\ref{tab:results-optimal-pruning}. 
PCKMT ($k$=8) equipped with pruning rate 30\% has the optimal performance, which obtains an improvement of 0.36 BLEU and 1.56x translation speed over the adaptive $k$NN-MT.

\begin{figure}[t]
    \centering
    \includegraphics[scale=0.3]{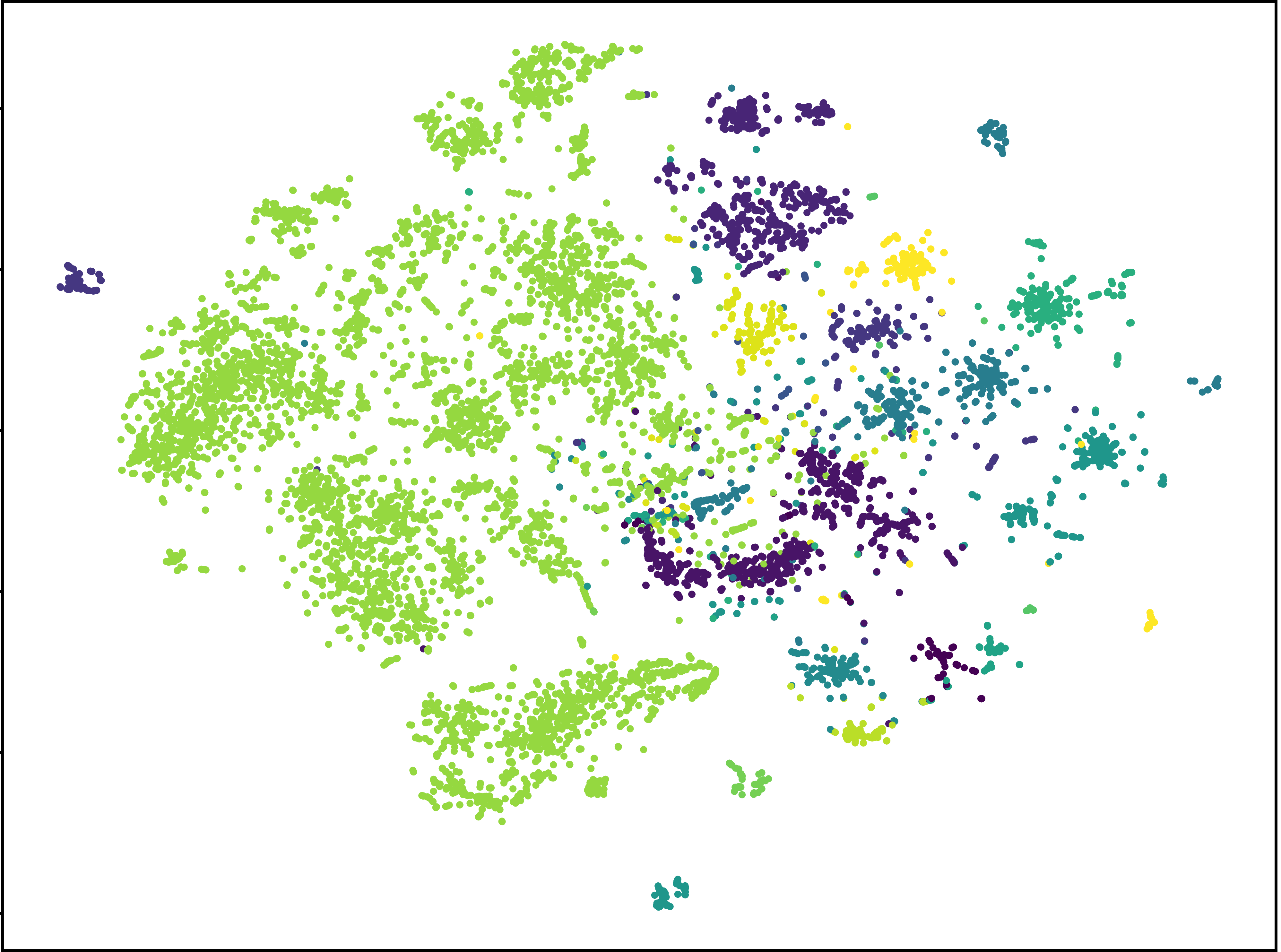}
    \includegraphics[scale=0.3]{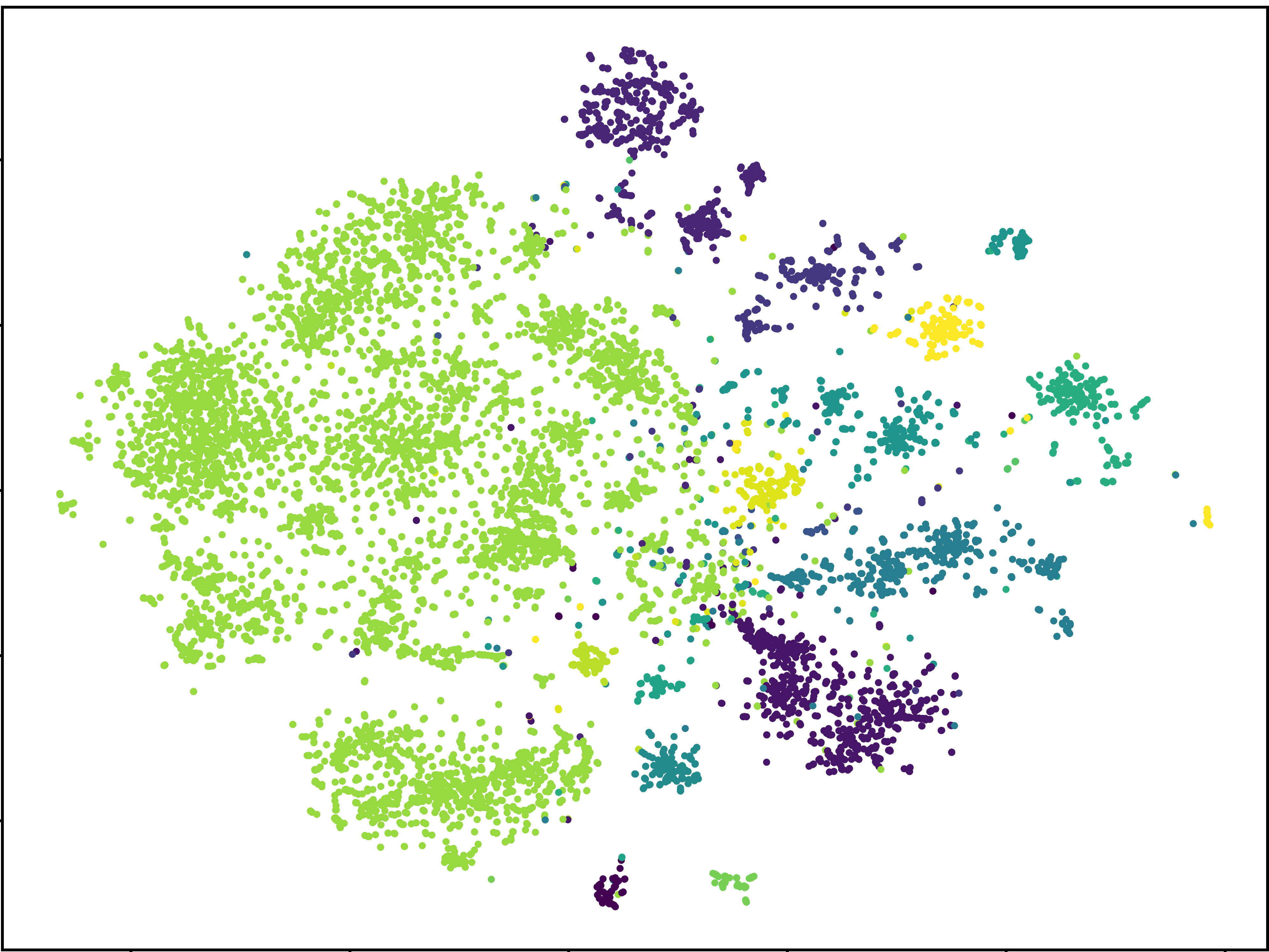}
    \caption{t-SNE visualization of the original (left) and compressed (right) semantic representations of 10 sampled tokens (one color for each token).}
    \label{fig:experiment-analysis-visualization}
\end{figure}
\textbf{Cluster Visualization.}
We visualize the IT domain datastore in Figure~\ref{fig:experiment-analysis-visualization} to verify our assumption that our Compact Network maps the original semantic representations to a separable distribution with less overlaps. Tokens represented by purple dots become more distinguishable with our method.

\section{Conclusion}

In this paper, we propose a cluster-based Compact Network for feature reduction in a contrastive learning manner to reduce 90+\% context feature dimension, and suggest a cluster-based pruning strategy to prune 10\%\textasciitilde40\% redundant keys in datastore while translation quality remains unchanged. 
Our proposed methods achieve better or comparable performance while reducing up to 57\% inference latency against the advanced non-parametric MT model on several benchmarks. 
For future work, it is promising to design effective feature reduction algorithms and pruning strategies based on more linguistic and cross-lingual information.

\section*{Acknowledgements}

Both Dexin Wang and Deyi Xiong were partially supported by the Natural Science Foundation of Tianjin (Grant No. 19JCZDJC31400). We would like to thank the anonymous reviewers for their insightful comments.

\newpage
\bibliography{acl}
\bibliographystyle{acl_natbib}

\appendix
\label{sec:appendix}
~
\newpage
~
\newpage

\section{Compact Feature Dimension}
\label{sec:compact-dimension-decision}

\begin{table}[h]
    \centering
    \begin{tabular}{c|ccc}
    \toprule
     Model & Dimension & Loss & BLEU\\
      \hline
     \multirow{4}{*}{CKMT*}
     & 128                   & 1.82 & 44.42 \\
     &  64                   & 1.84 & \textbf{44.49} \\
     &  32                   & 1.88 & 42.41 \\
     &  16                   & 2.03 & 39.49 \\
     \hline
     adaptive $k$NN-MT
     & -                     & 1.82 & 44.20 \\
    \bottomrule
    \end{tabular}
    \caption{BLEU scores of different compact feature dimensions on the IT domain validation set.}
    \label{tab:compact-dimension-decision}
\end{table}

The output dimension of the first FFN in $f_\alpha$ was empirically set as 4 times of the output dimension of the whole $f_\alpha$.
We then conducted greedy search on the IT domain validation set to obtain the optimal output dimension of $f_\alpha$ in our Compact Network. As shown in Table~\ref{tab:compact-dimension-decision}, 64d was the optimal setting superior to the adaptive $k$NN-MT.

\section{Hyper-parameters of Faiss}
\label{appendix:hyper-parameter}
We followed the default implementation setting of \citep{adaptiveknnmt}.
To be concrete, we adopted the FP16 precision to store keys.
The number of partition-based quantization centroids was set to 1024 while the number of selected invested lists at query time in the cell-probe method\footnote{https://github.com/facebookresearch/faiss/wiki/Faiss-indexes} was set to 32.
The size of per quantized vector in bytes was set to 64 except for CKMT with 16d/32d compact feature dimension in Table~\ref{tab:compact-dimension-decision} because the output size of the quantized vectors must be smaller than the size of the input features for quantization.

\section{Analysis on the Number of Parameters}
\begin{table}[h]
    \centering
    \begin{tabular}{c|c}
    \toprule
        Model   & Parameter  \\
       \hline 
        MT      &  269.7M   \\
        Adaptive $k$NN-MT   &  269.7M   \\
        CKMT*   &  270.0M \\
    \bottomrule
    \end{tabular}
    \caption{The number of parameters of different models. }
    \label{tab:analysis-parameters}
\end{table}

We compared the number of overall parameters of different systems in Table~\ref{tab:analysis-parameters}.
It can be seen that our optimal CKMT* only requires 0.1\% more parameters than the adaptive kNN-MT while it significantly decreases the latency. Hence CKMT* achieves an important speed-quality trade-off.

\section{Case Analysis}
\label{sec:case-analysis}
In this subsection, we analyze translations generated by different models on the test sets.

\begin{table*}[t]
    \centering
    \begin{tabular}{c|l|l|l}
    \toprule
     Source& Einfügen; 3D-Objekte & VolumeControl & Spielzug des schwarzen Spielers \\
     \hline
     
      Reference & \textbf{inserting}; 3-D objects & VolumeControl & Black's move\\
     \hline
     
     Base NMT
     & \textbf{Insert}; 3D objects
     & Volume control & Black Player's Move \\
     \hline
     
     adaptive $k$NN-MT
     & \textbf{pasting}; 3-D objects
     & api.op. & White's move   \\
     \hline
     
     CKMT*
     &  \textbf{inserting}; 3-D objects
     &  Volume Control & Black's move \\
     \hline
     
     PCKMT*
     & \textbf{inserting}; 3-D objects
     &  VolumeControl & Black's move \\
     \bottomrule
    \end{tabular}
    \caption{Translation examples generated by different models from the IT domain.}
    \label{tab:analysis-translation-example}
\end{table*}

From the translations generated by different models in Table~\ref{tab:analysis-translation-example}, it can be seen that CMKT* translates sentences more adequately especially for those containing ambiguous tokens because the Compact Network turns different tokens separable in the compressed semantic representation space.
In this way, CKMT* tends to predict accurate words that are in line with the meaning of the source sentence rather than tokens of high frequency (e.g., ``insert'' objects rather ``paste'' objects).
On the other hand, the adaptive $k$NN-MT translates ``VolumeControl'' as ``api.op'' by mistake while the base NMT model could correctly translate it, which suggests that the adaptive $k$NN-MT could surfer from noisy retrievals from the original semantic space.
It can also be seen that PCKMT* makes predictions without performance degradation compared to CKMT*,  although PCKMT* is equipped with a smaller datastore.

\end{document}